\documentclass[runningheads]{llncs}
\usepackage{amsmath,amssymb,amsfonts}
\usepackage{algorithmic}
\usepackage{graphicx}
\usepackage{booktabs}
\usepackage{subcaption}
\usepackage{hyperref}

\usepackage{float}
\usepackage[square,numbers]{natbib}
\usepackage[utf8]{inputenc}
\usepackage{xcolor}
\usepackage{longtable}
 
\usepackage{amsmath}

\DeclareMathOperator*{\argmin}{arg\,min}
\begin{document}
\section*{Acknowledgement}
This is a preprint of an article published in Nature Scientific Reports. The final authenticated version is available online at: \url{https://doi.org/10.1038/s41598-022-11826-0} (open access).

\title{Generalisation effects of predictive uncertainty estimation in deep learning for digital pathology \thanks{This work was supported by the Swedish e-Science Research Center and VINNOVA.}}
\titlerunning{Uncertainty and reliability of AI in digital pathology}
%
\author{Milda Pocevičiūtė\inst{1,2} \and
Gabriel Eilertsen \inst{1,2} \and
Sofia Jarkman \and \inst{2, 3}
Claes Lundström\inst{1,2,4}}
\authorrunning{M. Pocevičiūtė et al.}
%
\institute{Department of Science and Technology, Linköping University, Sweden\\
\email{\{milda.poceviciute, gabriel.eilertsen, sofia.jarkman, claes.lundstrom\}@liu.se} \and
Center for Medical Image Science and Visualization, Linköping University, Sweden \and Department of Clinical Pathology, and Department of Biomedical and Clinical Sciences, Linköping University, Linköping, Sweden \and
Sectra AB, Linköping, Sweden }
%




\maketitle              
\begin{abstract}
Deep learning (DL) has shown great potential in digital pathology applications. The robustness of a diagnostic DL-based solution is essential for safe clinical deployment. In this work we evaluate if adding uncertainty estimates for DL predictions in digital pathology could result in increased value for the clinical applications, by boosting the general predictive performance or by detecting mispredictions. We compare the effectiveness of model-integrated methods (MC dropout and Deep ensembles) with a model-agnostic approach (Test time augmentation, TTA). Moreover, four uncertainty metrics are compared. Our experiments focus on two domain shift scenarios: a shift to a different medical center and to an underrepresented subtype of cancer. Our results show that uncertainty estimates increase reliability by reducing a model's sensitivity to classification threshold selection as well as by detecting between 70\% and 90\% of the mispredictions done by the model. Overall, the deep ensembles method achieved the best performance closely followed by TTA.

\end{abstract}

\section{Introduction}
\label{sec:introduction}

The performance of deep learning (DL) has surpassed that of both human experts and other analytic methods on many prediction tasks in computer vision as well as other applications \cite{Laith2021, KOUMAKIS2020}. It has also shown great potential in pathology applications, such as breast cancer metastases detection \cite{Bejnordi2017, campanella2019, yun2019} and grading of prostate cancer \cite{strom2020, steiner2020, Pantanowitz2020, wang2021}. One remaining challenge for wide clinical adoption of DL in pathology, perhaps the most important one, is that the performance of neural networks (NNs) can deteriorate substantially due to domain shift \cite{Kouw2018AnIT}, i.e., differences in the distributions between the training data and the data for which prediction occurs. In digital pathology, differences of whole slide images (WSIs) are observed between different centres due to, for instance, medical protocols,  tissue preparation processes or scanner types used. 

The typical strategy to achieve high generalization capacity of a DL model is to ensure high diversity in the training data. In pathology, collecting data from several sources (care providers) is important. Another approach to reducing the domain shift error is to apply extensive augmentations during the training stage, which is an active research area in the pathology domain \cite{Tellez2019, Stacke2021}. Unfortunately, these methods are yet far from completely alleviating the performance drop due to domain shift. Thus, exploring more strategies to ensure a model's robustness when deployed is highly motivated.

The uncertainty of a prediction is a source of information that is typically not used in DL applications. There is, however, rationale indicating that it could be beneficial. By generating multiple, slightly varied, predictions as base for an uncertainty estimate, additional information about the model's sensitivity to changes is made available. This potential added value could be relevant for both the generalisation challenge and for boosting performance overall. In particular, we argue that the added dimension of uncertainty could be utilised as a building block for clinical workflows where pathologists and DL models interact, for instance for triaging WSIs or sorting WSI regions of interest.

In this paper, we explore a number of research questions that are central for understanding how uncertainty can contribute to robustness of DL for digital pathology:
\begin{itemize}
    \item Can uncertainty estimates add to the predictive capacity?
    \item Are uncertainty estimates effective for flagging potential prediction errors?
    \item Is the added value of uncertainty affected by domain shift?
    \item Are elaborate uncertainty methods more effective than simple ones?
    \item Must DL models be designed to provide uncertainty estimates, or can the same value be achieved with model-independent methods?
\end{itemize}

Our experiments explore two types of domain shift in histology of lymph nodes in breast cancer cases. One shift concerns data coming from two different care providers, in different countries. Another shift concerns the challenge of dealing with cancer subtypes, whether uncertainty estimation could be beneficial when there are rare conditions that may lack sufficient amount of training data.  Around 70\% of breast cancer cases are ductal carcinomas. The second most common subtype is lobular carcinomas which accounts to roughly 10\% of the positive cases. Furthermore, this subtype is often more challenging for a pathologist to detect due to its less obvious infiltrative patterns \cite{Li2003, Dossus2015}. Therefore, it is reasonable to assume that DL models not specifically trained for lobular carcinomas will have lower performance there. We use the ductal vs lobular carcinoma scenario as a proxy for the general case of subtypes that are under-represented in the training data.

With respect to uncertainty estimation methods, there is an important distinction between methods that influence the choice of NN architecture or training procedure, and methods that are independent of how the DL model was designed. The first category includes MC dropout \cite{Gal2016}, ensembles \cite{Lakshminarayanan2016} and other techniques  \cite{Camarasa2020, Linmans2020, Nair2020}. There are, however, important advantages of model-independent uncertainty estimation, such as Test time augmentations (TTA) \cite{Ayhan2018}.
One advantage is that constraints on model design may lead to suboptimal performance. Moreover, model independence opens the possibility to benefit from uncertainty estimates for any model -- also for the locked-down commercial solutions that are typically deployed in the clinic. Thus, studying the effectiveness of model-independent uncertainty estimation is of a particular interest. To the best of our knowledge, we are the first to compare the effectiveness of TTA, MC dropout, and ensembles on classification tasks in digital pathology. 

In this work, we train an NN classifier as a base for our evaluation of uncertainty methods. We contribute to the understanding of uncertainty and deep learning for digital pathology in four ways. First of all, we propose a way of combining the uncertainty measure with the softmax score in order to boost generalisability of the model. Secondly, we measure how well misclassified patches can be deteced by uncertainty methods. Thirdly, we compare the effectiveness of three uncertainty methods (Deep ensembles, MC dropout, and TTA) together with four different metrics utilising the multiple predictions -- three established measures (sample variance, entropy and mutual information) and our proposed metric (sample mean uncertainty). Finally, we investigate if uncertainty estimations generalise over a clinically realistic domain shift, and for mitigating the problem of a rare cancer subtype that is under-represented in the training data.

\section{Related work}
Uncertainty estimation is an important topic in deep learning research that holds potential in providing more calibrated predictions and increasing the robustness of NNs. The methods can be categorised based on what statistical theory they are grounded on: frequentist approaches, Bayesian neural networks (BNNs) and Bayesian approximations for standard NNs \cite{poceviciute2020}. The methods based on frequentist statistics commonly use ensembles \cite{Lakshminarayanan2016, mariet2021distilling}, bootstrapping \cite{Osband2016} and quantile regression \cite{Tagasovska2018}. BNNs are 
based on Bayesian Variational Inference and estimate the posterior distribution for a given task, and thus provide uncertainty distributions over parameters by design. However, currently their adaptation to the medical imaging domain is slow due to the higher computational costs of training and poor uncertainty estimation \cite{poceviciute2020,Wenzel2020}. 
There is also a more recent line of research showing that certain transformations of the softmax confidence score~\cite{Pearce2021}, or some modification to the network architecture~\cite{mukhoti2021}, may produce a reasonable estimation of uncertainty without any additional computations. To reflect this recent trend we include a comparison of direct uncertainty estimation from softmax score in all of our experiments, as well as an uncertainty estimator based on the sample mean over different network evaluations.

In deep learning applications within the medical domain, most research effort has been devoted to radiology, with MC dropout and Deep ensembles being two common methods compared in the literature. Nair et al.~\cite{Nair2020} showed that the MC dropout \cite{Gal2016} method can improve multiple sclerosis detection and segmentation. They evaluated the uncertainty measures by omitting a certain portion of the most uncertain predictions and comparing the effect on false positive and true positive rates. Kyono et al.~\cite{Kyono2020} evaluated if AI-assisted mammography triage could be safely implemented in the clinical workflow of breast cancer diagnosis. They estimated uncertainty by combining the MC dropout and TTA methods, and concluded that this approach could provide valuable assistance.

Within computational pathology, the most similar previous work is by Thagaard et al.~\cite{Thagaard2020}, which evaluated the deep ensembles \cite{Lakshminarayanan2016}, MC dropout \cite{Gal2016}, and mixup \cite{Zhang2017} methods for breast cancer metastases detection in lymph nodes. They trained an NN model for breast cancer metastasis detection and evaluated its performance in combination with the three uncertainty estimation methods on several levels of domain shifts: in-domain test data (same hospital, same organ), breast cancer metastases in lymph nodes from a different hospital,  colorectal cancer (different hospital and organ), and head and neck squamous cell carcinoma (different hospital, organ and sub-type of cancer) metastases to the lymph nodes. They found that Deep ensembles \cite{Lakshminarayanan2016} performed considerably better on most evaluation criteria except for detecting squamous cell carcinoma where mixup \cite{Zhang2017} showed better results. Similarly, Linmans et al.~\cite{Linmans2020} showed that uncertainties computed by Deep ensembles as well as a multi-head CNN~\cite{Lee2015} allowed for detection of out-of-distribution lymphoma in sentinel lymph nodes of breast cancer cases.

TTA in medical imaging has successfully been applied for segmentation tasks. Graham et al.~\cite{Graham2019} improved the performance of gland instance segmentation in colorectal cancer by incorporating TTA uncertainties into the NN system. Wang et al.~\cite{Wang2019} compared the potential gains from using MC dropout, TTA or a combination of both on segmentation performance of fetal brains and brain tumours from 2D and 3D magnetic resonance images. They found that the combination of the two methods achieved the best results.

In comparison with previous research efforts, our work brings novel contributions in several ways. This includes evaluating the model-agnostic TTA method for classification in pathology, and making the comparison to model-integrated methods. We introduce an approach to combine a model's softmax score with an uncertainty measure in order to improve the predictive performance. Moreover, we use a broader evaluation scheme for misprediction detection where all classification thresholds are considered instead of a single one. The broad scheme also includes evaluation of three uncertainty estimation methods using four different metrics, whereas previous work mostly has focused on the entropy metric. Finally, in all experiments we include a baseline based on the softmax score from one single model, in order to clearly measure the improvement that can be achieved by the added complexity of uncertainty estimation methods.

\section{Uncertainty estimation methods and metrics}
In this section we describe the three uncertainty estimation methods and the four uncertainty metrics that are evaluated in our experiments.

\subsection{Uncertainty estimation methods}
All of the methods have the same basic principle: to produce multiple predictions for each input. The variation within these predictions shows how uncertain the model is.   

\subsubsection{MC dropout} We are interested in computing posterior probability distribution $p(W| X, Y)$  over the NN weights $W$ given the input patches $X$ and corresponding ground true labels $Y$. This posterior is intractable, but it can be approximated using variational inference with some parameterised distribution $q^{*}(W)$ that minimises the Kullback-Leibler (KL) divergence:
\[q^{*}(W) = \argmin_{q(W)} KL(q(W) \| p(W| X, Y)).\]
Gal et al.~\cite{Gal2016} showed that minimising the cross-entropy loss of an NN with dropout layers, is equivalent to minimising the KL divergence above. Furthermore, the authors show that we can treat the samples obtained by multiple stochastic passes through an NN with the dropout enabled as an approximation of the model's uncertainty. Following Thagaard et al.~\cite{Thagaard2020}, we added a dropout layer with probability 0.5 in the NN before the logits. During test time, we activated the dropout layer with the same probability and ran 50 stochastic passes for each input.

\subsubsection{Deep ensembles} This is a method based on training $T$ identical NNs with different random seeds. During the inference, the $T$ predictions per input are aggregated for uncertainty estimation \cite{Lakshminarayanan2016}. Following previous work \cite{Thagaard2020, Linmans2020}, we set $T=5$.

\subsubsection{Test time augmentations} Each input is randomly augmented $T$ times before passing through the trained model. The uncertainty scores are computed from the $T$ predictions. Usually, the test time augmentations are identical to the ones applied during the training of the model \cite{Ayhan2018, Wang2019}. In our experiments we set $T = 50$ to match the number of forward passes in the MC dropout method. For a detailed description of the augmentations, refer to Section \ref{sec_networkdetails}.

\subsection{Uncertainty metrics}
Once we obtain the multiple predictions per input, we can compute an uncertainty metric. In this work we compared three well established metrics, sample variance, entropy and mutual information. In addition, we introduce the \emph{sample mean uncertainty} metric which is based on a probabilistic interpretation of the softmax score in a binary classification problem.

\subsubsection{Sample mean uncertainty} This metric is based on the mean of the samples generated by an uncertainty estimation method. We define sample mean uncertainty, $u_s$, as:
\[ u_s = 1 - 2(\overline{s} - 0.5)^2, \] 
where $\overline{s}$ is the average of softmax scores $s_i$ over $T$ predictions:
\[\overline{s} = \frac{1}{T} \sum_{i=1}^{T} s_i.\]
The value range of the measure is between 0 and 1, and assigns high value for patches that have the mean tumour softmax score around 0.5, indicating that they are potentially more uncertain. Low values are observed when the softmax scores are close to 0 or 1, implying high confidence in the corresponding binary classifications. The measure reflects the general dependence between softmax confidence and uncertainty.

Also, it shares characteristics with the estimator based on max predicted softmax probability for any class, which was evaluated in~\cite{Pearce2021}.

\subsubsection{Sample variance} This metric is derived by taking the variance across $T$ number of predictions per input produced by each of the uncertainty methods \cite{Nair2020}.

\subsubsection{Entropy} For a discrete random variable $X$, Shannon entropy quantifies the amount of uncertainty inherent in the random variable's outcomes. It is defined as \cite{shannon1948}:
\[H(X) = - \sum_{i} P(x_i) \log P(x_i),\]  
which we approximate for each input $i$ as \cite{Gal2016entropy}:

\begin{align*}
     H(\hat{y_i}| \mathbf{W}, \mathbf{D}) & \approx - \sum_{c=1}^{C}  \frac{1}{T} \sum_{t=1}^{T} \Bigg[ P(\hat{y_i} = c| W_t, D_t) \\
     & \cdot \log \left(\frac{1}{T} \sum_{t=1}^{T} P(\hat{y_i} = c| W_t, D_t) \right) \Bigg],
\end{align*}
where $T$ is the number of predictions per input generated by an uncertainty estimation method, $C$ is the number of classes in our data, $\mathbf{D}$ is the dataset, $\hat{y_i}\text{, } i \in |\mathbf{D}|$ is a prediction by the classifier, and $\mathbf{W}$ are the parameters of the classifier. We refer to this metric as 'entropy'.

\subsubsection{Mutual information (MI)} The MI metric was first defined by Shannon \cite{shannon1948}. It measures how much information we gain for each input by observing the samples produced by an uncertainty estimation method. It is approximated by \cite{Gal2016entropy}:
\begin{align*}
    MI(\hat{y_i}, \mathbf{W}| \mathbf{D}) \approx H(\hat{y_i}| \mathbf{W}, \mathbf{D}) - E \left[ H(\hat{y_i}| W_t, D_t) \right] ,
\end{align*}
where $ H(\hat{y_i}| \mathbf{W}, \mathbf{D})$ is the entropy of expected predictions. $E \left[ H(\hat{y_i}| W_t, D_t) \right]$ is the expected entropy of model predictions across the samples generated by an uncertainty estimation method which can be approximated as \cite{Gal2016entropy}:
\begin{align*}
    E \left[ H(\hat{y_i}| W_t, D_t) \right] & \approx - \sum_{c=1}^{C}  \frac{1}{T} \sum_{t=1}^{T} \Bigg[ P(\hat{y_i} = c| W_t, D_t) \\ & \cdot 
     log (P(\hat{y_i} = c| W_t, D_t) ) \Bigg]
\end{align*}
\section{Implementation details and data}
In this section we describe the NN algorithms that we trained for the classification task, the training procedure, and the datasets used for training and evaluation of the uncertainty methods and metrics.

\subsection{Network training}
\label{sec_networkdetails}

We trained five Resnet18 models \cite{He2016} with He initialisation \cite{He2015} and a dropout layer (with probability 0.5) \cite{Srivastava2014} before the logits with five different random seeds. The data augmentations during the training as well as the testing time were based on the work of Tellez et al.~\cite{Tellez2019}. That is, on each input we applied horizontal flip with probability 0.5, 90 degrees rotations, scaling factor between 0.8 and 1.2, HSV colour augmentation by adjusting hue and saturation intensity ratios between  [-0.1, 0.1], brightness intensity ration: [0.65, 1.35], contrast intensity ratio: [0.5, 1.5]. We also applied additive Gaussian noise and Gaussian blur, both with $\sigma \in [0.0, 0.1]$. 

Each training epoch consisted of 131 072 patches sampled from the training WSIs with equal number of tumour and healthy patches. We used ADAM optimiser with $\beta_1 = 0.9$, $\beta_2 = 0.999$, initial learning rate of $0.01$ with learning rate decay of 0.1 applied when the validation accuracy was not improving for 4 epochs. The models were trained until convergence with the maximum limit of 100 epochs.

\subsection{Datasets}

\begin{table}
\caption{Information about the datasets used in the evaluation of the model and the uncertainty methods.}
\label{tab_datasets}
\setlength{\tabcolsep}{3pt}
\begin{tabular}{p{47pt}|p{45pt}p{20pt}p{30pt}p{30pt}p{35pt}}
\hline
                    & Country       & Total WSIs & Positive WSIs   & Patches  & Cancer types  \\ \hline
Camelyon16 (test) & Netherlands     &  129  & 49 & 40 940   & Not available
\\ \hline
BRLN data       & Sweden            & 114 & 57    &  39 354  & Ductal,  lobular  \\ \hline
Lobular data    & Sweden            & 71 & 14    &  6 960  & Lobular  \\ \hline
Ductal data     & Sweden            & 96 & 39     &  6 960  & Ductal  \\ \hline
\end{tabular}

\end{table}

In-domain data in this project is the Camelyon16 dataset \cite{Litjens2018} which contains 399 whole-slide images (WSI) of hematoxylin and eosin (H\&E) stained lymph node sections collected in two medical centres in the Netherlands. The slides were scanned with the 3DHistech Pannoramic Flash II 250 and Hamamatsu NanoZoomer-XR C12000-01 scanners. 270 WSIs from Camelyon16 dataset were used for the training and validation of Resnet18 models while the official test set consisting of 129 WSIs was kept for the in-domain performance evaluation. In our experiments, 'Camelyon16 data' refers to the Camelyon16 test set unless otherwise noted.

Our out-of-domain data is 114 H\&E stained WSIs of lymph node sections from a medical center in Sweden annotated by a resident pathologist with 4 years of experience aided with immunostained slides. This is a subset of the larger AIDA BRLN dataset \cite{jarkman019}, which was scanned by Aperio ScanScope AT and Hamamatsu NanoZoomer scanners (XR, S360, and S60). We refer to it as BRLN.

Table \ref{tab_datasets} lists the four datasets of patches extracted from Camelyon16 and BRLN that were used in our experiments. These datasets were only used for the evaluation. In BRLN data, we have two cancer subtypes: lobular and ductal carcinomas. 
In order to study uncertainty effects on generalisation to the lobular cancer subtype, we created two subsets of BRLN data which we call Lobular and Ductal data. They consist of 3480 tumour patches of each cancer subtype and the same 3480 healthy patches. 

\subsection{Evaluation metrics}
We evaluate our results based on area under the curve (AUC) of receiver operating characteristic (ROC) and precision recall (PR). 
ROC-AUC is the most common metric used to evaluate the performance of a binary classifier~\cite{Bekkar2013} and also in uncertainty evaluation in digital pathology \cite{Thagaard2020, Graham2019, Linmans2020}. ROC-AUC captures the trade-off between the true positive rate (TPR), also known as recall, and false positive rate (FPR), also known as 1 - specificity:

\[TPR = \text{Recall} = \frac{TP}{TP + FN} \]
\[FPR = 1 - \text{Specificity} = \frac{FP}{TN + FP}\]


The PR curve plots precision against recall where precision is the fraction of positive predictions that are truly positive \cite{Powers2011}:
\[ \text{Precision} = \frac{TP}{TP + FP}\] 

In addition to the AUC measures that aggregate performance across all classification thresholds, we are interested in examining in detail how performance of methods and metrics varies for different choices of classification thresholds. For this comparison, we look at accuracy:

\[\text{Accuracy} = \frac{TP + TN}{TP + FP + TN + FN} \]





\section{Results}

\begin{table}
\caption{PR and ROC AUC values based on softmax scores (single model) for each of the 4 datasets.} 
\label{tab_gen_results}

\resizebox{0.7\columnwidth}{!}{%
\begin{tabular}{l|llll }
\hline
            & Camelyon16 &  BRLN & Lobular & Ductal \\ \hline
 ROC-AUC    & 0.975       & 0.968 & 0.899   & 0.982   \\
 PR-AUC     & 0.981       & 0.977 & 0.928   & 0.987  \\ \hline
\end{tabular}
}
\end{table}

\subsection{Basis for the experiments}
\subsubsection{Performance of classifier}
In order to draw any meaningful conclusions on evaluation of uncertainty methods and metrics, we first need to ensure that the base classifier has a reasonable performance. 

Table \ref{tab_gen_results} shows ROC-AUC and  PR-AUC on the four datasets. 
Overall, the achieved performance of 0.975 ROC-AUC (0.981 PR-AUC) on in-domain Camelyon16 data indicates that the Resnet18 was a sufficient classifier to perform this detection task. The  drop of below 1\% in ROC-AUC (and PR-AUC) between Camelyon16 and BRLN data sets is consistent with other work observing that a well-trained model should suffer relatively small decrease in performance under domain shift arising from different medical centers \cite{campanella2019}.  

\subsubsection{Increased error for lobular carcinoma}
\label{sec:error_lobular}
Investigating the model's performance on cancer subtypes within BRLN data, we found that it exhibits a substantially worse result on the lobular carcinoma: 0.889 ROC AUC (0.928 PR AUC) compared to the 0.982 ROC AUC (0.987 PR AUC) on the ductal cancer subtype (see table \ref{tab_gen_results}). This result confirms that there indeed is a domain shift effect due to tumour type, in line with our assumptions.

\subsection{Boosting metastases detection}
\label{sec:threshold}

 \begin{figure}[t]
    \begin{minipage}[b]{1\linewidth}
      \def\cpar{\hss\egroup\line\bgroup\hss}
      \centering
      \centerline{\includegraphics[width=10.0cm]{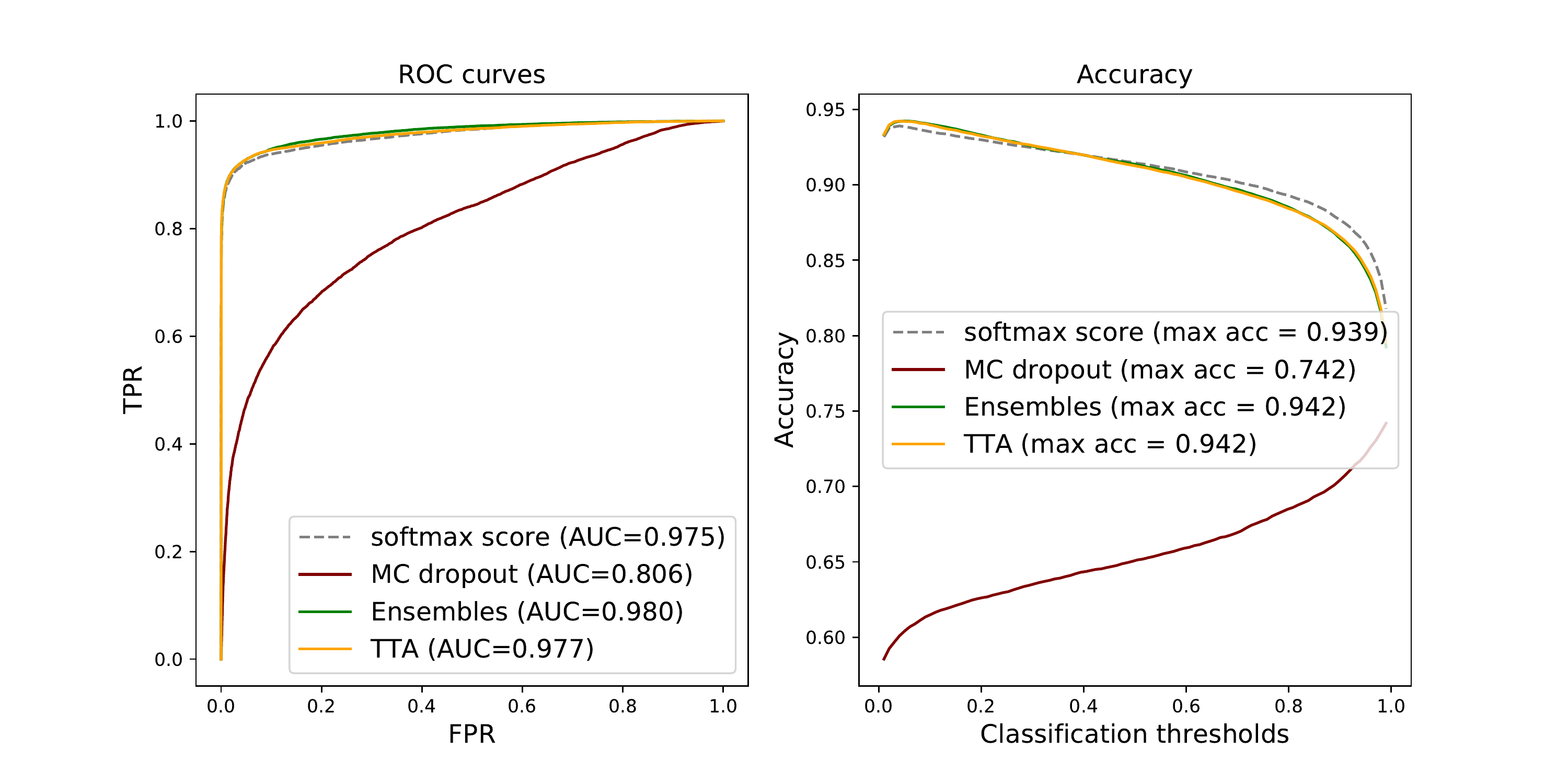}}
    \end{minipage}
    \centering
    \caption{Tumour metastases detection on Camelyon16 data: ROC curves and accuracy of using softmax tumour score from a single NN vs averages of softmax tumour scores (per input) produced by the uncertainty estimation methods.}
    \label{fig_softmax_means_cam16}
\end{figure} 

 \begin{figure}[t]
    \begin{minipage}[b]{1\linewidth}
      \def\cpar{\hss\egroup\line\bgroup\hss}
      \centering
      \centerline{\includegraphics[width=10.0cm]{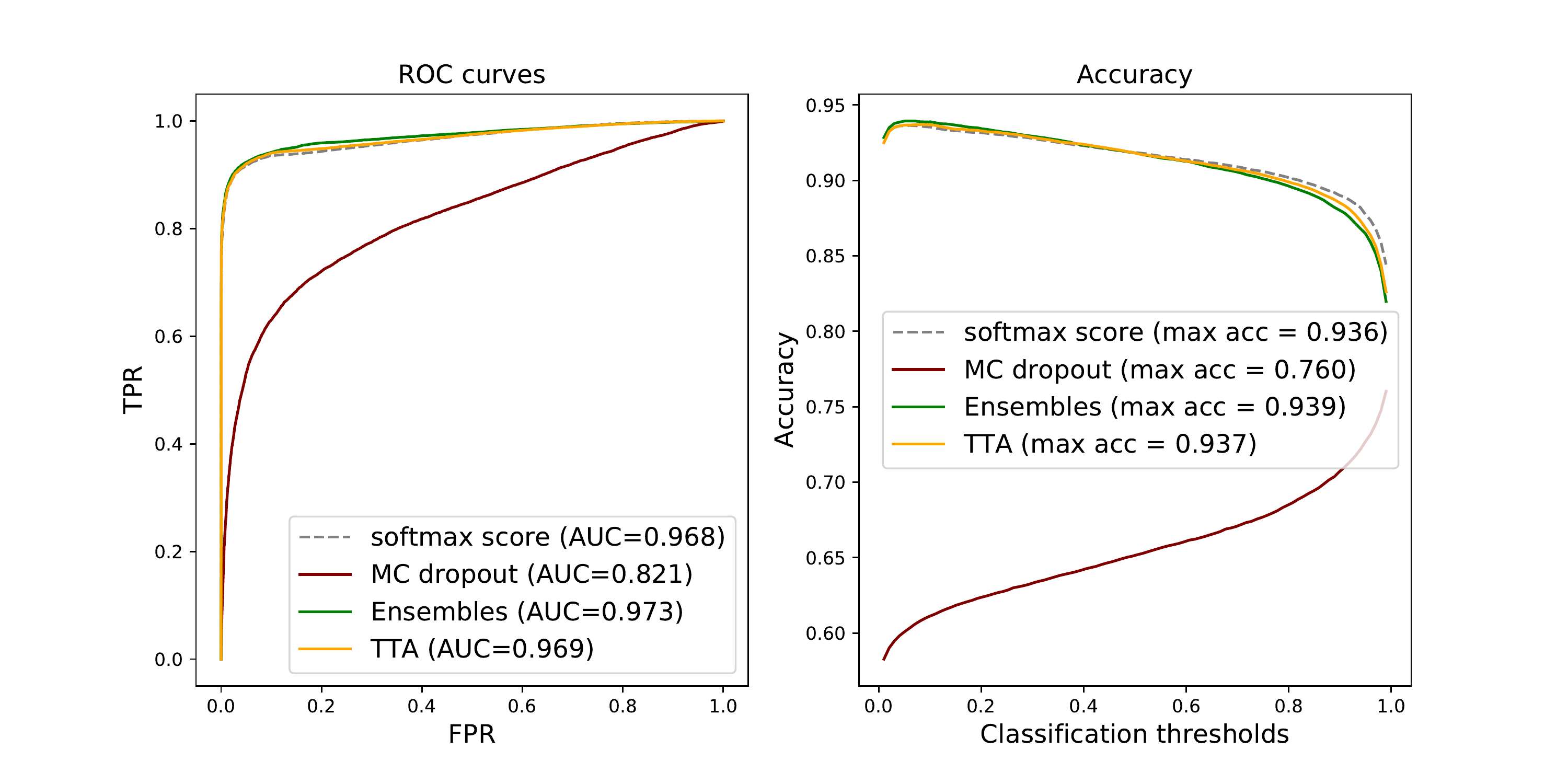}}
    \end{minipage}
    \centering
    \caption{Tumour metastases detection on BRLN data: ROC curves and accuracy of using softmax tumour score from a single NN vs averages of softmax tumour scores (per input) produced by the uncertainty estimation methods.}
    \label{fig_softmax_means_brln}
\end{figure}

\begin{figure}
    \centering
    \begin{subfigure}[b]{0.48\linewidth}
      \includegraphics[width=\linewidth,trim=4mm 4mm 4mm 4mm,clip]{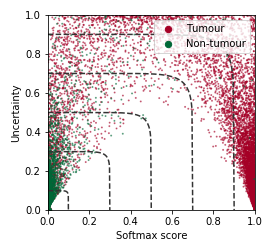}
      \caption{Ensemble, entropy}
      \label{fig:scatter_ensemble}
    \end{subfigure}
    \begin{subfigure}[b]{0.48\linewidth}
      \includegraphics[width=\linewidth,trim=4mm 4mm 4mm 4mm,clip]{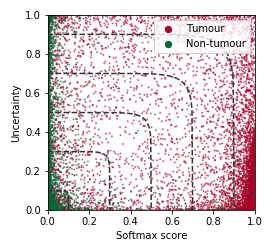}
      \caption{MC dropout, entropy}
      \label{fig:scatter_mcdrop}
    \end{subfigure}
    \caption{Relation between softmax confidence and estimated uncertainty, for two different uncertainty methods. The points show the testset from Camelyon16, with colors encoding ground truth class labels. The dashed lines illustrate the 2D threshold used for classification based on both softmax and uncertainty, for a set of different threshold values.}
    \label{fig:scatter}
\end{figure}

\begin{table}
\centering
\caption{ROC-AUCs on Camelyon16 and BRLN data: combining softmax tumour score from a single NN with uncertainty estimates. Softmax score refers to using the softmax output alone.}
\label{tab_roc_boost}
\resizebox{0.7\columnwidth}{!}{
\begin{tabular}{@{}lcccc@{}}
\hline
\toprule

 \textit{ROC-AUCs} \\
            & MC dropout & Ensembles  & TTA  & Softmax score   \\ 
\hline
 \textit{Sample mean uncertainty}\\
    Camelyon16  & 0.908  & 0.979  & 0.976   &   0.975 \\
    BRLN        & 0.913 & 0.971  & 0.968   & 0.968  \\
\midrule
 \textit{Sample variance}\\
    Camelyon16  & 0.962  &  0.975 &  0.975  &    0.975 \\
    BRLN     & 0.955 & 0.968  & 0.968   & 0.968  \\
  \midrule
 \textit{Entropy}\\
    Camelyon16  & 0.924  &  0.980 & 0.977   &    0.975 \\
    BRLN    & 0.924 & 0.972  &  0.969  & 0.968  \\
  \midrule
   \textit{Mutual information}\\
    Camelyon16  & 0.954  &  0.978 &  0.976  &    0.975 \\
    BRLN    & 0.947 & 0.970  & 0.968   &  0.968 \\
  \bottomrule    
  \hline
\end{tabular} }
\end{table}

 \begin{figure}[t]
    \begin{minipage}[b]{1\linewidth}
      \def\cpar{\hss\egroup\line\bgroup\hss}
      \centering
      \centerline{\includegraphics[width=9.0cm]{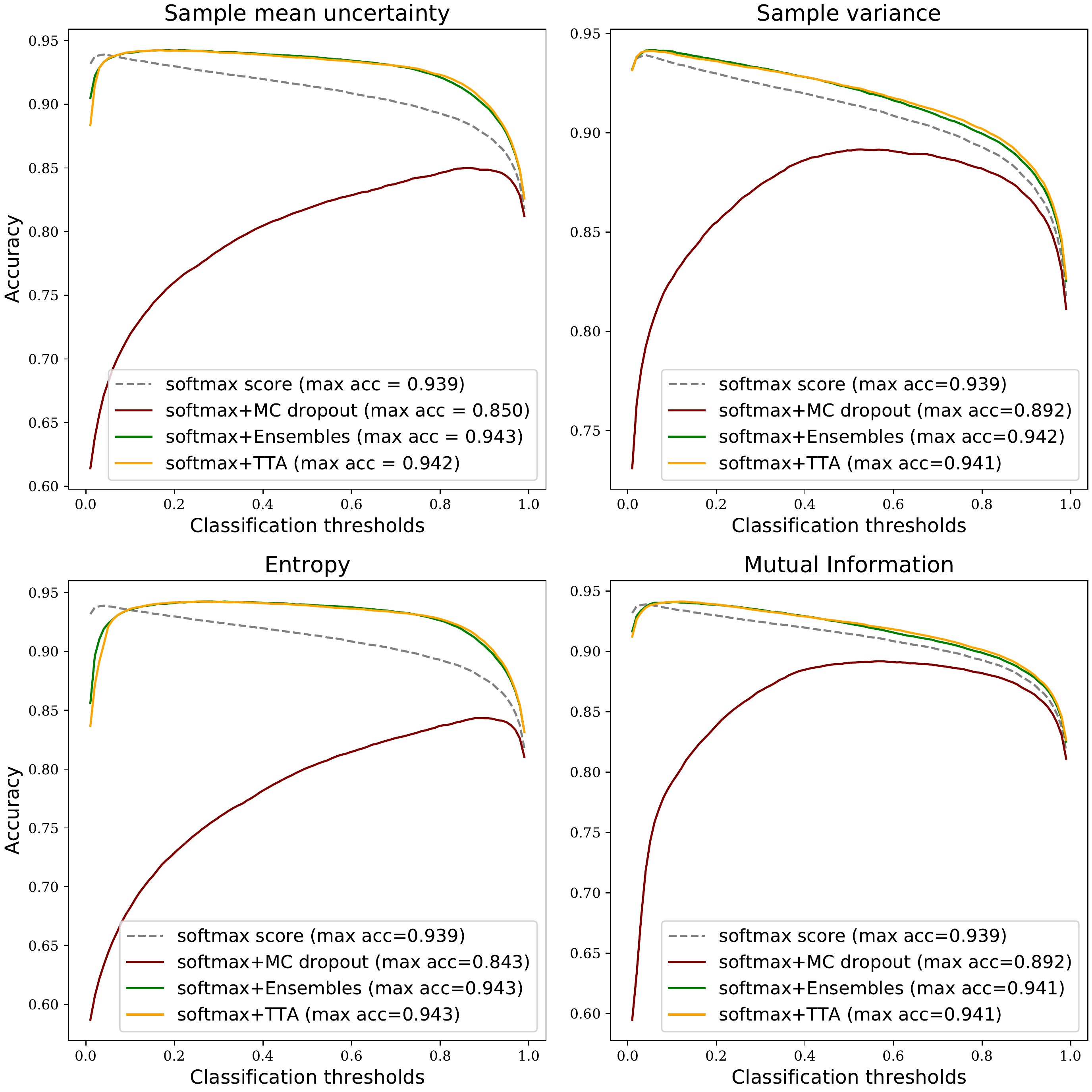}}
    \end{minipage}
    \centering
    \caption{Prediction accuracy on Camelyon16 data when combining softmax tumour score from a single NN with uncertainty estimates.}
    \label{fig_acc_boost_cam16}
\end{figure}


 \begin{figure}[t]
    \begin{minipage}[b]{1\linewidth}
      \def\cpar{\hss\egroup\line\bgroup\hss}
      \centering
      \centerline{\includegraphics[width=9.0cm]{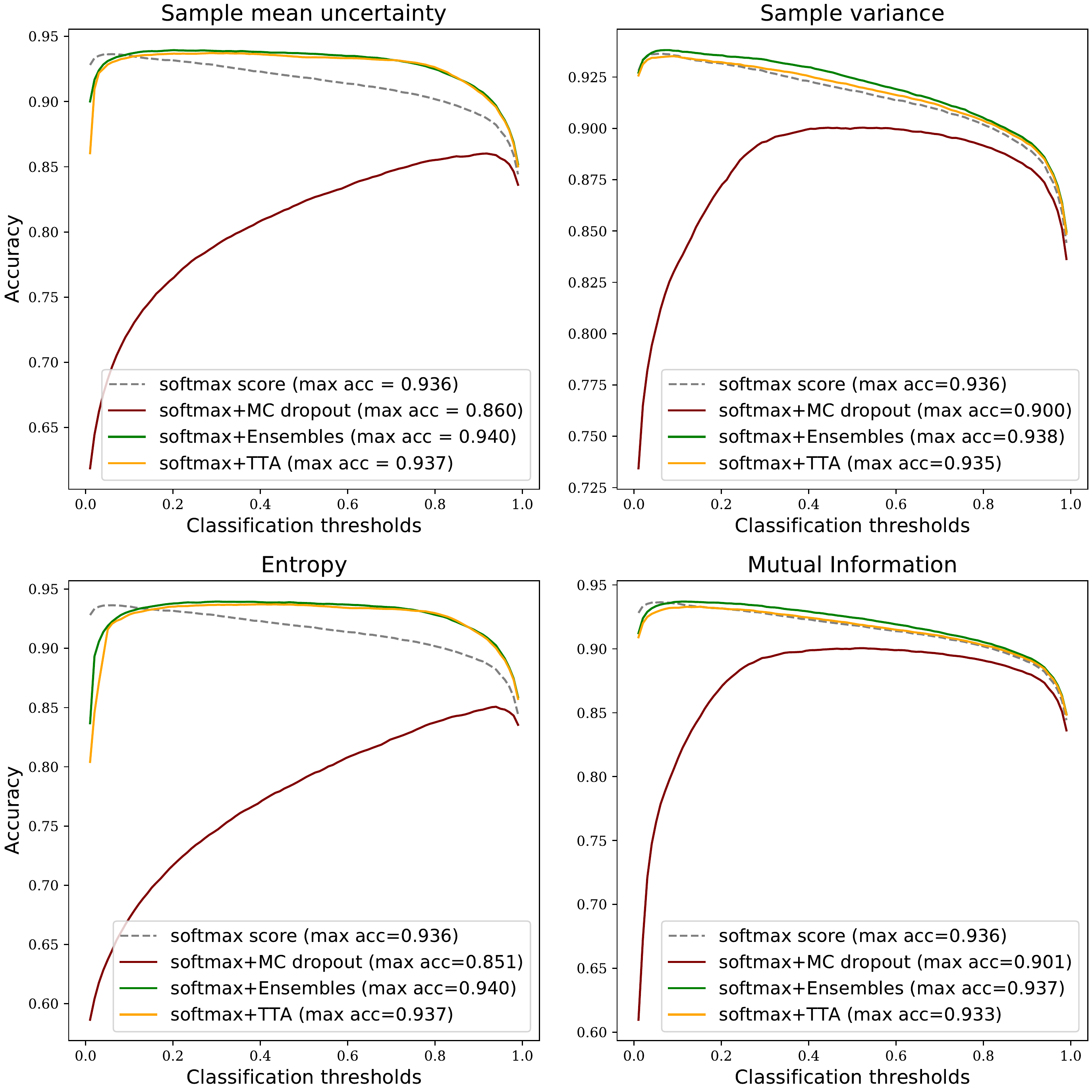}}
    \end{minipage}
    \centering
    \caption{Prediction accuracy on BRLN data when combining softmax tumour score from a single NN with uncertainty estimates.}
    \label{fig_acc_boost_brln}
\end{figure}

Given the multiple predictions provided by three different methods (MC dropout, ensembles, and TTA), the most straightforward method for boosting predictive performance is to utilise traditional ensemble techniques. The most common one is to average the softmax output over the different inference runs/models/augmentations. The results are demonstrated in Fig.~\ref{fig_softmax_means_cam16} and \ref{fig_softmax_means_brln} for Camelyon16 and BRLN, respectively, and compared to using a single prediction as baseline. The results show a consistent but small improvement in terms of ROC-AUC for the deep ensembles and TTA. This is also reflected by the accuracy curve, demonstrating how the averaging improves the results by a small margin ($\sim$0.3 percentage points) if calibrated with the optimal threshold, for both Camelyon16 and BRLN.

Another strategy for boosting predictive performance is to consider uncertainty estimation and softmax output from a single network as separate entities. Although the uncertainty methods also make use of different combinations of the softmax score, it is interesting to investigate if this approach holds benefits over traditional techniques.
We do this by turning the classification task into a two-dimensional thresholding problem, with the softmax score and the uncertainty measure as two separate dimensions. 
To test this strategy, our experiments aim to include uncertainty for reducing the number of false negatives. This can be achieved by rejecting negative predictions that have a higher uncertainty, simply changing their prediction to positive. We do this by using a 2D decision boundary:
\begin{equation*}
    f(u,s) = \left( \left( \frac{u}{P_u} \right)^y + s^y \right)^{\frac{1}{y}},
\end{equation*}
where $u$ is the uncertainty measure, $s$ is the softmax score for the tumour (positive) class from one single NN. The factor $P_u$ is used to normalise the range of uncertainties, and we define this to be the 99th percentile of the uncertainty value range in the data. Based on a specified threshold $t$, the prediction is positive for $f(u,s)>t$, otherwise negative. The curve $f(u,s)=t$ intersects the axes at $t$, and the exponent $y$ can be used to control its shape, from circular for $t=2$ towards square for large $t$. For all experiments we use $t=10$. Fig.~\ref{fig:scatter} illustrates the 2D space spanned by softmax score and uncertainty estimation, for two different methods, with corresponding 2D decision boundaries, $f(u,s)=t$, for a selection of different $t$. 


Table \ref{tab_roc_boost}  summarises the ROC-AUC results on Camelyon16 and BRLN data, for different combinations of uncertainty metrics and methods. We can see that MC dropout is the only method that, independently of metric and data set, achieves worse ROC-AUC scores than the softmax score from a single NN.
TTA and deep ensembles exhibit nearly identical performance for each computed metric. The uncertainty-including methods consistently perform at par or better than the baseline of using the softmax score from a single NN, but the improvement is small, below 1 percentage point in terms of ROC-AUC. 

Although the ROC-AUC results are similar compared to the traditional ensemble technique (Fig.~\ref{fig_softmax_means_cam16} and~\ref{fig_softmax_means_brln}), another aspect of robustness is how the performance of methods and metrics varies across the range of classification thresholds.
In Fig.~\ref{fig_acc_boost_cam16} we can see that when using the sample mean or entropy uncertainty, the shape of the accuracy versus classification threshold curves are considerably different for Camelyon16 data. Instead of a narrow range of peak accuracy, we get high performance over a broader range of thresholds. This indicates that embedding uncertainty information can lessen the sensitivity for how the operating point of the prediction is set, which is one part of the generalisation challenge.
Importantly, this finding holds true also under domain shift (Fig.~\ref{fig_acc_boost_brln}). 



\subsection{Misprediction detection}
\label{sec_mispredictions}

\begin{figure}[t]
    \begin{subfigure}[b]{0.48\linewidth}
      \centering
      \includegraphics[scale=0.22]{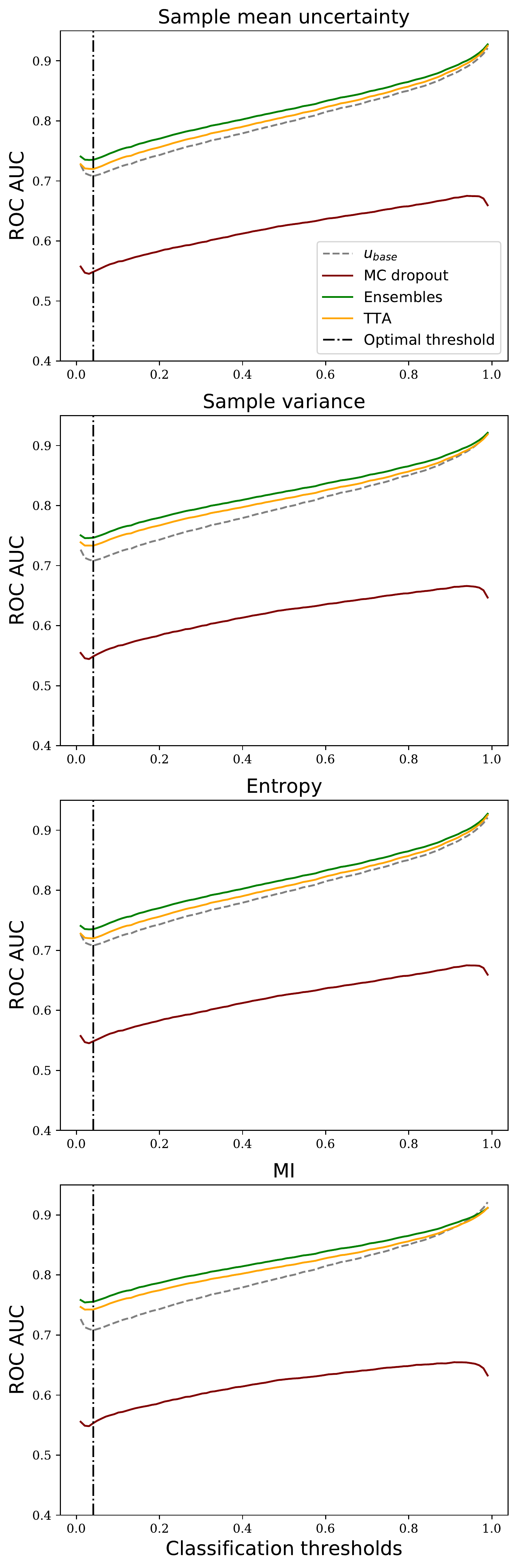}
      \caption{Camelyon16 data}
    \end{subfigure}
    \hfill
    \begin{subfigure}[b]{0.48\linewidth} 
      \centering
      \includegraphics[scale=0.22]{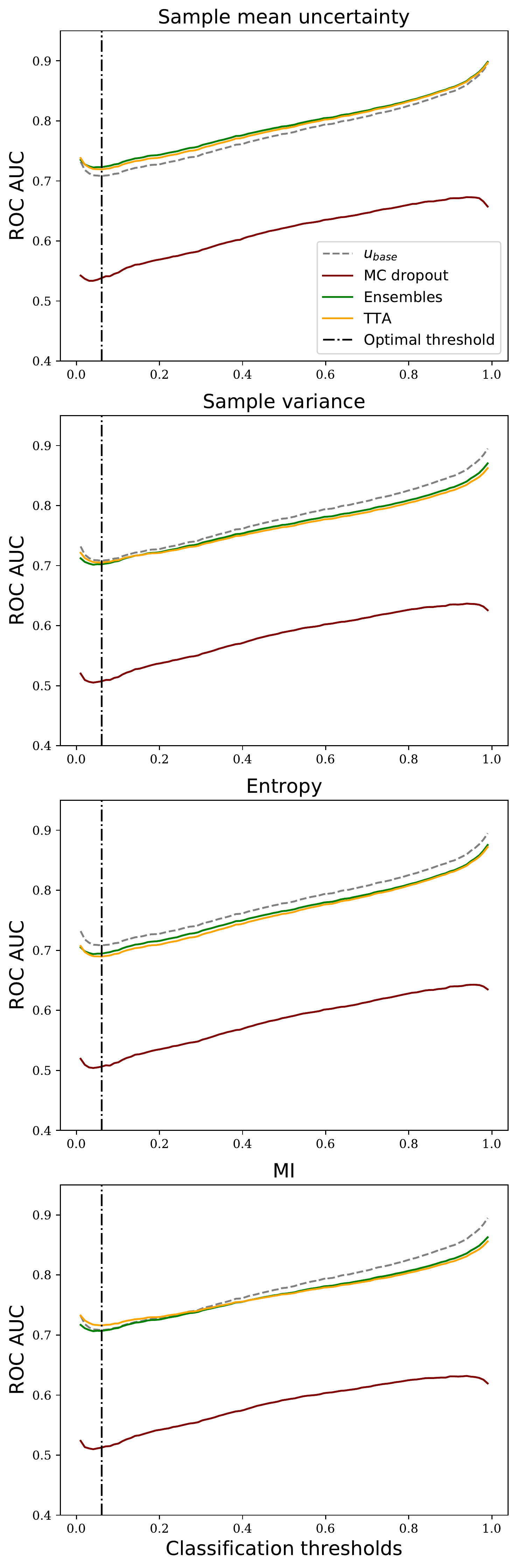}
      \caption{BRLN data}
    \end{subfigure}
    \centering
    \caption{ROC-AUCs of misprediction detection on Camelyon16 (in-domain) and BRLN (domain shift) data sets for different thresholds used to differentiate between tumour and non-tumour predictions. The softmax-based baseline uncertainty is the same in all plots.}
    \label{fig_mispred_results}
\end{figure}

\begin{table}
\centering
\caption{\textbf{Camelyon16 data:} ROC-AUCs of misprediction detection for varying classification thresholds.}
\label{tab_sr_aucs_cam16}
\resizebox{0.7\columnwidth}{!}{
\begin{tabular}{@{}lccccc@{}}
\hline
\toprule
          & MC dropout & Ensembles  & TTA  & Baseline & Classification \\ 
        &  &  &   & & Accuracy \\ 
\hline
 \textit{Sample mean}\\
    threshold 0.1  & 0.566  & 0.751  & 0.736   &   0.723 & 0.935\\
    threshold 0.5  & 0.626 & 0.818   & 0.807   & 0.798 & 0.914 \\
    threshold 0.9  & 0.672 &  \textbf{0.891} &  0.885  &  0.879 & 0.875 \\
\midrule
 \textit{Sample variance}\\
    threshold 0.1  &  0.567 & 0.762  & 0.749  &  0.723 & 0.935\\
    threshold 0.5  &  0.627  &  0.824 &  0.812 & 0.798  & 0.914 \\
    threshold 0.9  & 0.665  & 0.889  &  0.882  & 0.879  & 0.875 \\
  \midrule
     \textit{Entropy}\\
    threshold 0.1  &  0.566  & 0.751  &  0.736 &  0.723 & 0.935 \\
    threshold 0.5  & 0.626  &  0.818 & 0.807  &  0.798 & 0.914 \\
    threshold 0.9  &  0.672 & \textbf{0.891} &  0.885 &  0.879 & 0.875 \\
  \midrule
   \textit{Mutual information}\\
    threshold 0.1 &  0.571  & \textbf{0.770} & 0.757  & 0.723  & 0.935 \\
    threshold 0.5 &  0.626 &   \textbf{0.828} & 0.816  &  0.798 & 0.914 \\
    threshold 0.9  & 0.655 & 0.886  &  0.879 &  0.879 & 0.875 \\
  \bottomrule    
  \hline
\end{tabular} }
\end{table}

\begin{table}
\centering
\caption{\textbf{BRLN data:} ROC AUCs of misprediction detection for varying classification thresholds.}
\label{tab_sr_aucs_brln}
\resizebox{0.7\columnwidth}{!}{
\begin{tabular}{@{}lccccc@{}}
\hline
\toprule
          & MC dropout & Ensembles  & TTA  & Baseline & Classification \\ 
        &  &  &   & & Accuracy \\ 
\hline
\midrule
 \textit{Sample mean uncertainty}\\
    threshold 0.1  &  0.547 &  \textbf{0.728} &  0.724 & 0.712  & 0.935 \\
    threshold 0.5  &  0.622 &  \textbf{0.791}  & 0.788  & 0.779  &  0.918 \\
    threshold 0.9  &  0.671 &   \textbf{0.857} & 0.856  &  0.850 & 0.889 \\
  \midrule
 \textit{Sample variance}\\
    threshold 0.1  &   0.514 &  0.707 &  0.710 & 0.712  & 0.935 \\
    threshold 0.5  & 0.590  & 0.768  & 0.764  & 0.779  &  0.918 \\
    threshold 0.9  & 0.635  &  0.831  &  0.826 & 0.850  & 0.889 \\
  \midrule
     \textit{Entropy}\\
    threshold 0.1  &   0.513 &  0.700 & 0.694  & 0.712  & 0.935 \\
    threshold 0.5  &  0.588 &  0.766 & 0.761  & 0.779  &  0.918 \\
    threshold 0.9  &  0.640 &  0.833  &  0.831  &  0.850 & 0.889 \\
  \midrule
   \textit{Mutual information}\\
    threshold 0.1 & 0.519  & 0.712  & 0.719  & 0.712  & 0.935 \\
    threshold 0.5 & 0.592  & 0.769  &  0.768  & 0.779  &  0.918 \\
    threshold 0.9  & 0.631  & 0.827  &  0.823 &  0.850 & 0.889 \\
  \bottomrule    
  \hline
\end{tabular} }
\end{table}

In addition to embedding uncertainty information in the prediction, a straightforward application of the uncertainty estimates is in misprediction detection. Performance for this task also provides a general idea about the capacity of the methods to boost robustness in a deployed diagnostic tool.

We compare the three uncertainty estimation methods incorporating multiple predictions with a baseline uncertainty $u_\text{base}$ derived from a single softmax value:
\[ u_\text{base} = 1 - 2(s - 0.5)^2, \]
where $s$ refers to the softmax output for the tumour class of a single NN. The baseline captures the general correlation between softmax score and uncertainty, where uncertainty is maximal at $0.5$ and decreases towards $0$ and $1$, as seen in Fig.~\ref{fig:scatter_ensemble}.

\subsubsection{Evaluating uncertainty methods}
From the plots in Fig.~\ref{fig_mispred_results} showing the ROC-AUC performance for the misprediction detection task, we observe the same tendency as in the experiment of boosting the general performance: ensembles and TTA are substantially better than MC dropout. In fact, the latter performs worse than the baseline independently of the chosen metric or classification threshold.

In table \ref{tab_sr_aucs_cam16}, we see that the highest result for all classification thresholds was achieved by ensembles method. TTA performance is midway between the baseline and the ensembles. Comparing with table \ref{tab_sr_aucs_brln}, we see that domain shift affects misprediction performance in a negative way. Under the domain shift, only ensembles and TTA with sample mean uncertainty consistently achieve improvements over the baseline, whereas other combinations are at par with or below the baseline (see also Fig.~\ref{fig_mispred_results}b). 

An interesting observation is that there is a trade-off between how good the uncertainty methods are at misprediction detection and how well the NN performs on its primary task of cancer metastases detection. For higher threshold values, the predictive accuracy of the NN decreases, but the misclassification detection effectiveness increases (Fig.~\ref{fig_mispred_results}). This may suggest that uncertainty estimation is more beneficial for models with weaker predictive performance. 

\subsubsection{Evaluating uncertainty metrics}

In the experiments we also compared the four uncertainty metrics. In  Fig.~\ref{fig_mispred_results}a, we observe that on the in-domain data all metrics achieve similar good performance compared to the baseline, when computed from TTA or deep ensembles predictions. From table \ref{tab_sr_aucs_cam16}, MI emerges as the best performing metric, closely followed by the other three. 

Sample variance, entropy and MI metric do not generalise well under the domain shift. From Fig.~\ref{fig_mispred_results}b and \ref{tab_sr_aucs_brln}, we see that sample mean uncertainty is the only metric that performs better than the baseline independently of the classification threshold on the BRLN data (for ensembles and TTA). 


\subsection{Uncertainty and lobular carcinoma}
\label{sec:lobular}

 \begin{figure}[t]
    \begin{minipage}[b]{1\linewidth}
      \def\cpar{\hss\egroup\line\bgroup\hss}
      \centering
      \centerline{\includegraphics[width=6.0cm]{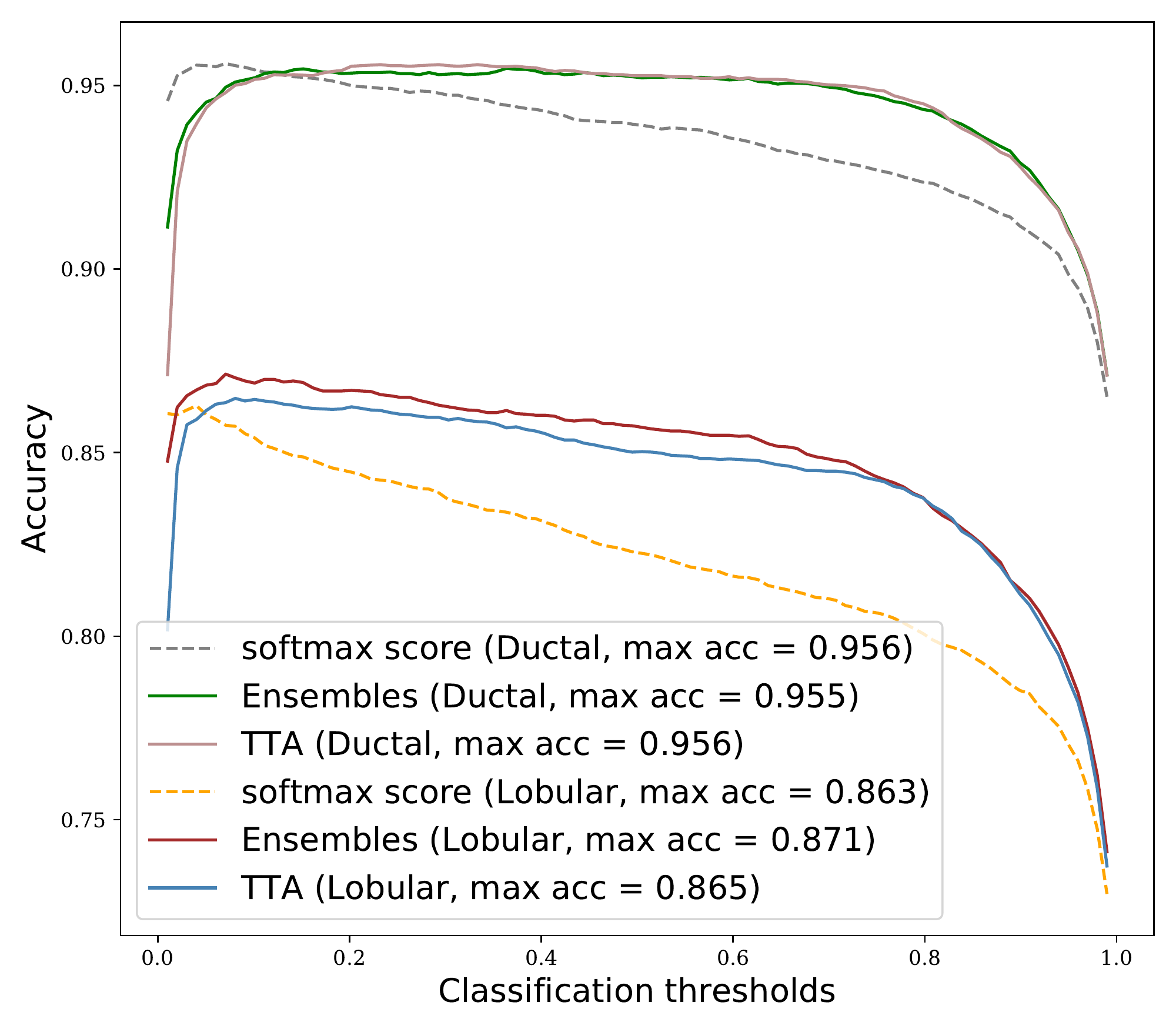}}
    \end{minipage}
    \centering
    \caption{Prediction accuracy on Lobular and Ductal data when combining softmax tumour score from a single NN with sample means uncertainty estimated by ensembles and TTA methods.}
    \label{fig_acc_boost_lobular_ductal}
\end{figure}

\begin{table}
\centering
\caption{Tumour metastases detection on Lobular and Ductal data: ROC AUCs of combination of sample means uncertainty and the softmax score.}
\label{tab_boost_lobular_ductal}
\resizebox{0.7\columnwidth}{!}{
\begin{tabular}{@{}lcccc@{}}
\hline
\toprule
\textit{ROC-AUC} \\
         & Ensembles  & TTA  & Baseline \\ 
\hline
     \textit{Sample mean uncertainty}\\
    Lobular data  & 0.908 &  0.899 &  0.899 \\
    Ductal data  & 0.984  & 0.983 &  0.982 \\
  \bottomrule    
  \hline
\end{tabular}}
\end{table}

 \begin{figure}[t]
    \begin{minipage}[b]{\linewidth}
      \def\cpar{\hss\egroup\line\bgroup\hss}
      \centering
      \centerline{\includegraphics[width=6.0cm]{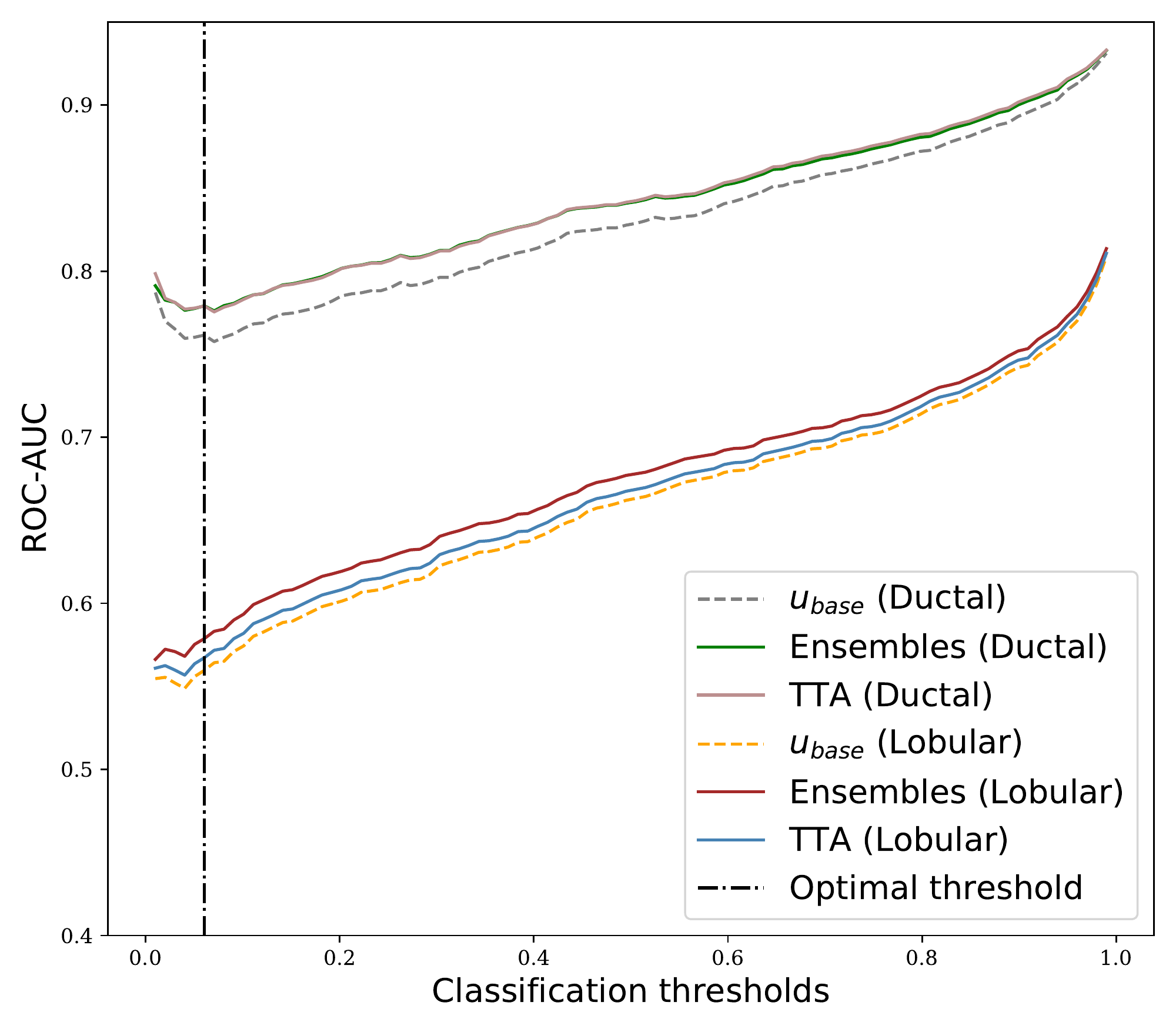}}
    \end{minipage}
    \centering
    \caption{ROC-AUCs of misprediction detection by sample means uncertainty from ensembles and TTA (for varying classification threshold). Baseline is computed from softmax score of a single NN.}
    \label{fig_mispreds_lobular_ductal}
\end{figure} 

\begin{table}
\centering
\caption{ROC-AUCs of misprediction detection by sample means uncertainty computed from ensembles and TTA methods.}
\label{tab_mispred_aucs_lobular_ductal}
\resizebox{0.7\columnwidth}{!}{
\begin{tabular}{@{}lcccc@{}}
\hline
\toprule
\textit{ROC-AUC} \\
         & Ensembles  & TTA  & Baseline & Classification \\ 
         &   & & & Accuracy \\ 
\hline
    \textit{Lobular data }\\
    threshold 0.1 & \textbf{0.593}   &  0.581  & 0.574  & 0.854 \\
    threshold 0.5 &  \textbf{0.678 }  & 0.669   & 0.663  & 0.823 \\
    threshold 0.9 &  0.754   & \textbf{0.756 }  & 0.744  & 0.784 \\
  \midrule
    \textit{Ductal data}\\
    threshold 0.1 &  \textbf{0.784}   &  0.783  & 0.766  & 0.954 \\
    threshold 0.5 &  \textbf{0.842}   &\textbf{0.842}   &  0.829  & 0.939 \\
    threshold 0.9 &  0.903    &  \textbf{0.904 } &  0.896  & 0.910 \\
    \bottomrule    
  \hline
\end{tabular}}
\end{table}

Now we turn to evaluating if uncertainty measures may contribute to boosting the performance on a rare type of data, in our case: lobular carcinoma. 
For this experiment, we focus on the consistent good performers in previous experiments: the sample mean uncertainty metric combined with the ensembles and TTA uncertainty estimation methods.

\subsubsection{Uncertainty for boosting the tumour metastases detection} In table \ref{tab_boost_lobular_ductal}, we see similar results as for the entire BRLN dataset: the ROC-AUCs improve marginally by combining the uncertainty with the softmax score, slightly more improvement for the lobular data. Fig.~\ref{fig_acc_boost_lobular_ductal} shows the previously noted effect of a flattened accuracy curve, where the accuracy increase for suboptimal thresholds is more pronounced for the lobular data set. 

\subsubsection{Uncertainty for misprediction detection} From Fig.~\ref{fig_mispreds_lobular_ductal} we conclude that all methods are substantially better at detecting mispredictions on the ductal cancer subtype than the lobular, meaning that this type of domain shift also has a negative effect on misprediction performance. For the optimal classification threshold, the misprediction performance on lobular data is not much better than a random guess.
From table \ref{tab_mispred_aucs_lobular_ductal} we see that the improvement from the baseline for the best performing uncertainty estimation method is similar on ductal and lobular data. 

\section{Discussion}
The main research question was whether uncertainty estimates can add to the predictive capacity of DL in digital pathology. The results show that uncertainty indeed adds value if good measures and metrics are chosen. The predictive performance can be slightly increased, but a perhaps more important benefit is a lessened sensitivity to the choice of classification threshold -- mitigating the infamous AI 'brittleness'. Uncertainty used for misprediction detection is valuable in the sense that performance is far above a random guess. The results also show, however, that the added value of introducing uncertainty over softmax probability is quite limited and it is an open question whether these benefits would make a substantial difference when employed in a full DL solution in a clinical setting.


Drilling down into detailed results, it is clear from the experiments that MC dropout is the least suitable method as the variability in its output has minimal value for boosting the general NN's performance directly or via misprediction detection. This is also apparent from inspecting the relation between softmax confidence and MC dropout uncertainty in Fig.~\ref{fig:scatter_mcdrop}, which show little correlation. 
In contrast, the TTA and deep ensemble methods outperformed the baseline on both evaluation tasks for most metrics. 
While deep ensembles exhibited the best performance, the difference to TTA was often negligible. Thus, if the flexibility offered by using a model-agnostic method is important in the scenario considered, TTA could be preferred.

Interestingly, the gains of using ensembles or TTA were larger for the classification thresholds corresponding to high accuracy, at least for the most well-performing metrics. Furthermore, our results demonstrate that misprediction detection is easier when classification is poor. This underlines that misprediction detection should not be considered in isolation, instead the interplay with classification accuracy should always be considered.

The choice of uncertainty metric is not trivial. In our experiments, entropy and sample mean uncertainty can be said to have achieved the best results overall, but the differences are small between all metrics. It is a somewhat surprising result that a mean aggregation performs at par with a metric taking variance into account.

In the out-of-domain experiments we saw diminishing performance gains from all combinations of uncertainty estimation methods and metrics. While this is consistent with previous work \cite{Thagaard2020}, it is discouraging, as the foremost objective of these approaches is to mitigate the generalisation problem. It seems that the variation of model output is not that different between in-domain and out-of-domain pathology data. In fact, only the sample mean uncertainty sustains a better performance than the simple softmax-based baseline in the out-of-domain case, and the baseline showed the least drop in performance due to domain shift. This is somewhat surprising, as we would have expected the softmax baseline to be more sensitive to domain shift. The reason is likely both that we deal with a smaller, clinically realistic, domain shift and that softmax can behave better than expected in out-of-domain situations~\cite{Pearce2021}. The upside of this result is that even a simple uncertainty measure can exhibit a reasonable performance on misprediction detection.


In the study of detecting mispredictions within a data subtype that is underrepresented in the training set (lobular carcinoma), we observed that uncertainty methods and the baseline are much less effective at this compared to the abundant data subtype (ductal carcinoma). The performance gains from using ensemble and TTA uncertainty estimation had larger margin for the classification thresholds corresponding to the highest accuracy, but less than on the in-domain data.

One of the limitations of this work is that we worked with patches extracted from WSIs. This was essential to investigate the basic properties of the uncertainty in digital pathology, but a study on how this translates to WSI level decisions is necessary. Furthermore, we focused on breast cancer metastases detection in the lymph nodes. More studies should be carried out to confirm that the results hold in other digital pathology applications. Of particular interest is to study prediction tasks with lower accuracy, where our results indicate that the added value of uncertainty may be greater than in this work. Regarding TTA, there may be other types of augmentations that are better suited to the specific objective of estimation of predictive uncertainty. There are also other method parameter options that could be relevant to evaluate. The dropout probability chosen for MC dropout may, for instance, not be optimal for our ResNet18 architecture, but we argue (also in light of previous work) that it is unlikely that the MC dropout performance then would surpass the other methods. Finally, some more extensive tuning of the uncertainty estimation methods or using a combination of several uncertainty methods could be evaluated in future work.

\section{Conclusion}

We conclude that the evaluated uncertainty methods and metrics perform well on in-domain data but are affected by the domain shift due to new medical center as well as the underrepresented subtypes of data in the training set. The softmax score of the target NN can be transformed to provide an uncertainty measure which is less affected by the domain shift than the more established methods. The associated computational costs and NN design constraints indicate that the use of softmax score transformation is an appealing alternative to the uncertainty estimation methods. 

\bibliographystyle{splncs04}
\bibliography{bibliography.bib}


\end{document}